# Semantic Segmentation from Limited Training Data

A. Milan[1,3], T. Pham[1,3], K. Vijay[1,3], D. Morrison[1,2], A.W. Tow[1,2], L. Liu[3],
J. Erskine[1,2], R. Grinover[1,2], A. Gurman[1,2], T. Hunn[1,2], N. Kelly-Boxall[1,2], D. Lee[1,2],
M. McTaggart[1,2], G. Rallos[1,2], A. Razjigaev[1,2], T. Rowntree[1,3], T. Shen[1,2] R. Smith[1,2],
S. Wade-McCue[1,2], Z. Zhuang[1,4], C. Lehnert[2], G. Lin[1,3], I. Reid[1,3], P. Corke[1,2], and J. Leitner[1,2]

*Abstract*— We present our approach for robotic perception in cluttered scenes that led to winning the recent Amazon Robotics Challenge (ARC) 2017. Next to small objects with shiny and transparent surfaces, the biggest challenge of the 2017 competition was the introduction of *unseen* categories. In contrast to traditional approaches which require large collections of annotated data and many hours of training, the task here was to obtain a robust perception pipeline with only few minutes of data acquisition and training time. To that end, we present two strategies that we explored. One is a deep metric learning approach that works in three separate steps: semantic-agnostic boundary detection, patch classification and pixel-wise voting. The other is a fully-supervised semantic segmentation approach with efficient dataset collection. We conduct an extensive analysis of the two methods on our ARC 2017 dataset. Interestingly, only few examples of each class are sufficient to fine-tune even very deep convolutional neural networks for this specific task.

## I. INTRODUCTION

Robotic solutions have been utilised in the industry for many years. However, their use is typically restricted to known and structured environments with pre-defined actions. Examples include parts assembly in the automotive industry or produce sorting in the grocery sector. Manipulating individual items in a large warehouse is a more complex task for machines and to date remains unsolved. Even though the environment can be controlled to a certain degree, e.g. the storage system can be designed in a certain way to facilitate recognition, the sheer number of items to be handled poses non-trivial challenges. In addition, the items are often placed in narrow bins to save space, thus partial or even full occlusion must be addressed from both the perception and manipulation side.

Quantitatively evaluating robotic solutions is not a trivial task. To allow for a fair comparison and also to advance the state of the art in autonomous warehouse manipulation, in 2017, Amazon organised the third Amazon Robotics Challenge (ARC), previously known as the Amazon Picking Challenge. Sixteen qualified teams competed on the tasks of stowing and picking common objects. The first one consisted of emptying a crowded tote into a self-designed

This research was supported by the Australian Research Council Centre of Excellence for Robotic Vision (ACRV) (project number CE140100016). The participation at the ARC was supported by Amazon Robotics LLC. Contact: trung.pham@adelaide.edu.au

[1] Authors are with the Australian Centre for Robotic Vision (ACRV).
[2] Authors are with the Queensland University of Technology (QUT).
[3] Authors are with the University of Adelaide.
[4] Authors are with the Australian National University (ANU).

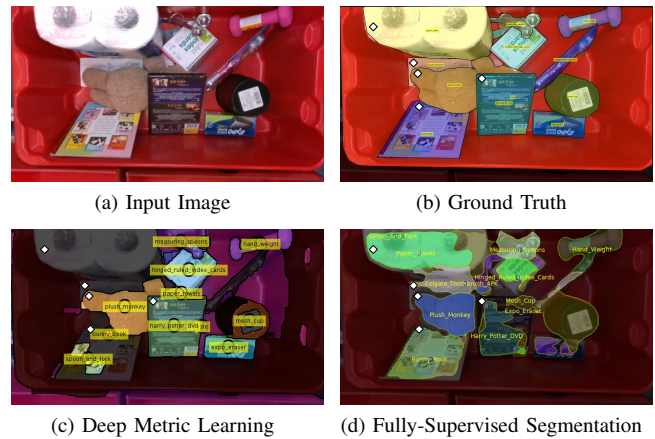

(a) Input Image  (b) Ground Truth
(c) Deep Metric Learning  (d) Fully-Supervised Segmentation

Fig. 1. An example of semantic segmentation for object picking. (a) and (b) show an input image and its ground-truth. (c) and (d) are the results of the Deep Metric Learning and fully-supervised semantic segmentation approach, respectively. The tote segment is not visualised. White diamonds mark unseen items.

storage system, while the latter required the robot to pack three different orders into different sized cardboard boxes, consisting of 2, 3, and 5 items, respectively. The two tasks were combined in the final round. Both the number of items in the bins as well as their appearance made the perceptions task slightly more difficult compared to previous editions. However, the biggest change was in the replacement of the training objects by new unseen ones, which were presented to the participants only 45 minutes prior to the start of each task. These conditions require very robust solutions that are not fully over-fitted to the training set, but also models that can be quickly adapted to new categories.

In this work we present two different approaches that address both challenges, which we examined during the development phase. Our initial strategy was to entirely bypass the need for data collection and model training for new categories. To that end, we first learn a feature embedding [1] which transforms image patches of the same object into low-dimensional points that are close-by, and patches of different objects into points that a far apart in the feature space. Classification can then be performed by a simple nearest neighbour look up. The patches are generated by a class-agnostic RGB-D segmentation [2]. To obtain the final segmentation map, a pixel-wise voting scheme is performed on all segments.

As our second approach, we explore a fully-supervised se-mantic segmentation solution based on deep neural networks. It is common practice to finetune existing models to specific tasks [3], [4], [5], [6]. While this is fairly straightforward in cases where all categories are known and defined beforehand, it is not such a common strategy for tasks where the amount of available training data is very limited. Nonetheless, it turns out that our choice of a deep architecture, which is RefineNet [4], is able to adapt to new categories using only very few training examples. To collect these examples, we follow an efficient data collection strategy, capturing seven views of each new object and using semi-automatic segmentation. The CNN is then fine-tuned for a few minutes only and reaches a level of performance that is sufficiently accurate for autonomous object manipulation. Fig. 1 illustrates typical semantic segmentation results using the two approaches. In summary, our main contributions are as follows.

- We present the full perception approach of the winning team of the Amazon Robotics Challenge 2017.
- We examine a deep metric learning approach that does not require any on-the-fly training and can easily handle unseen categories.
- We adapt RefineNet, a deep convolutional neural network for the specific task of autonomous picking and demonstrate how such deep architectures can be fine-tuned with very few training examples.
- We conduct a series of experiments and quantitatively compare the two orthogonal approaches.
- We compile and release a manually labelled dataset for 40 training objects and 16 validation objects, annotated with pixel-wise semantic segmentation masks.

After reviewing related work in the next section, we first provide a general overview of our system and hardware used in Sec. III and then present both our perception strategies in Sec. IV and Sec. V, respectively. Finally, we discuss our experiments in Sec. VI.

## II. RELATED WORK

The body of literature on robotic vision is vast. Here we only concentrate on relevant work concerning object perception for bin picking applications, in particular focusing on the specific task of stowing and picking items within the Amazon Challenge series.

Perception is considered the most challenging part of the problem, as clearly indicated in the survey by Correll et al. [7], which was conducted for the first Amazon Picking Challenge (APC) in 2015. RBO [8], the winning team of the first competition approached the object segmentation problem using one RGB-D sensor and without employing any deep learning techniques. Rather, manually designed colour and geometry features were used to learn feature histograms and then estimate per-pixel posterior probabilities for each object class. The maximum across all classes then yields the most likely object for each image region. A 3D bounding box is then fitted to the hypothesis to determine the grasp point. Their perception pipeline is described in detail in [9]. Team MIT [10] used depth measurements to register previously scanned object models to the obtained point cloud in order to determine the object pose. In their 2016 approach [11], the scene was measured from 15 to 18 views to produce a very dense point cloud. Similar to their previous solution, the objects were then registered to the point cloud using a modified ICP algorithm. In addition, semantic segmentation of each of the views was done using a VGG-style architecture [12]. To train the deep model, the team required over 100,000 labelled training images, which were collected and annotated semi-automatically using background subtraction. This strategy is reminiscent to our data collection approach described in Sec.V-A, but in our case we require 3 orders of magnitude fewer data samples.

Contrary to the above approach, the winning solution in 2016 designed by Team Delft [13] did not rely on pixel-wise segmentation prediction, but rather used Faster-RCNN [14], an object detection algorithm. The bounding boxes only provide a coarse estimate of the object extensions. To obtain more accurate object localisation, Hernandez et al. [13] use depth cameras to register known object geometry to the point cloud, while rejecting potentially wrong measurements using heuristics. Note that our second approach presented in this paper does not rely on depth sensors but rather uses RGB information only. Team NimbRo [3], [15] integrated both object detection and semantic segmentation into their vision system. The former is based on DenseCap [16] and adapted for the task at hand, while the latter uses the OverFeat architecture [17], [6] to learn strong features. Both solutions incorporate depth information obtained from three sources: two depth sensors and a stereo RGB setup.

Our first approach is inspired by the recent work on deep metric learning [18], [1], [19], [20] which shows that the feature embedding learned on seen categories generalises well to unseen object classes. The global loss [1] uses the batch statistics to overcome the limitations of the triplet network. Lifted structure embedding [18] uses a structured loss objective on the pairwise distance matrix to improve the embedding, while [19] employs a global clustering quality metric in the loss function. In this work, we use [1] to learn a feature embedding and then use a nearest-neighbour classifier to recover the class of the input image. The second strategy that we develop falls into the category of fully-convolutional networks (FCN). Since they were first introduced by Long et al. FCN [21] for semantic segmentation, FCNs have been explored for this specific task in various ways. DeepLab [22] uses so-called atrous (or dilated) convolutions to prevent excessive downscaling of the original input image. The Pyramid Scene Parsing Network (PSPNet) [23] introduces a special pooling module to aggregate context from different levels of the image pyramid. Mask R-CNN [24] is an extension of the widely known Faster R-CNN detection framework [14] that includes an instance-level segmentation prediction. In this work, we build on RefineNet [4], a multi-path refinement network that offers a good trade off between model complexity and training efficiency. Interestingly, it can be fine-tuned to new categories using only few training samples. This approach is described in more detail in Sec. V.

## III. SYSTEM OVERVIEW

The main focus of this paper is on the perception system behind Cartman [25], the robot that won the 2017 Amazon Robotics Challenge. For completeness, we briefly describe the hardware as well as the integration of our perception solution into the system, but refer the reader to [25] for further details.

Cartman is a Cartesian manipulator composed of a three-axis gantry that moves an end-effector comprised of a gripper and a sucker. The objective of Cartman is to move items between a storage system, stow tote and packing boxes, positioned atop the floor below. The gripper and sucker are positioned opposite one another; one of the wrist motors enables switching between the two tools. Cartman uses two Intel RealSense SR300 cameras for perception. The primary camera is mounted to the underside of Cartman's wrist, enabling vision into the storage system, stow tote or packing boxes beneath. A secondary camera is mounted to the side of Cartman, looking across the top of the storage system towards a red sheet. The secondary camera allows for photographing a grasped item on a plain background. Scales are positioned beneath the storage system and stow tote to provide additional feedback through measurement of weight deltas pre and post grasp. Scales allow errors in perception to be captured by ensuring the weight change measured matches the weight of the intended item.

Within the software system behind Cartman, semantic segmentation provides the link between a raw image of the scene and grasp synthesis. Grasp synthesis provides a pose at which Cartman should position its end-effector to enable a grasp to be performed. Grasp synthesis assumes that all visible items in the scene have been segmented. With an item segment provided, a hierarchy of approaches, each relying on a different level of available depth information, is performed to select an appropriate grasp pose. More details of the grasp synthesis approach can be found in [25], [26].

## IV. DEEP METRIC LEARNING

Deep learning based object recognition approaches have shown great success [27], [12], [28], [14], [22]. Such deep models, however, usually require a large collection of ground-truth data for training. Recognising unseen objects with only a few sample images available remains a difficult task. Here we attempt a deep metric learning approach to recognising unseen objects without requiring re-training the system. The idea is to learn a feature embedding via a deep neural network that transforms images of different views of the same object into similar low-dimensional features.

### A. Geometric Segmentation

Object contour detection and object segmentation in images are dual problems. One can easily obtain segmentation if a contour is available, and vice versa. As contours are independent from semantic classes, contour detection approaches are able to generalise to unseen categories and novel scenes without any re-training. This property is well suited for the picking challenge where unknown object categories are introduced during the competition.

The adoption of Convolutional Neural Networks (CNNs) has made significant progresses in many computer vision tasks such as object detection [14] and semantic segmentation [21]. Modern approaches to low-level tasks such as contour detection also achieve impressive performance [29], [2]. Here we resort to the convolutional oriented boundary (COB) network [2] to predict object boundaries directly from RGB-D images. In particular, the COB network predicts multi-scale multi-orientation contour probability maps, which are then combined into a single Ultrametric Contour Map (UCM) [30] — a hierarchical representation of the image. Figure 2 illustrates multiple segmentation hypotheses at multiple scales. One drawback of this approach is that it is not obvious which level will yield the optimal segmentation. We bypass this issue by passing all regions at all levels to the object classification step (see Section IV-B), and then use pixel-voting (see Section IV-C) to arrive at the final semantic segmentation for the entire image. In practice, we remove regions that are either above or below certain sizes, i.e., regions bigger than $50\%$ or smaller than $0.1\%$ of the image size are rejected.

As our robot is equipped with a color and a depth camera for perception, we use RGB-D images for boundary detection. We adopt the COB model trained on NYU RGB-D dataset [31], which somewhat resembles our bin picking data. No fine-tuning is performed. This COB model takes an RGB and HHA [32] images, as input. The HHA image encodes depth, height above ground and angle to gravity and is normalised to $[0, 255]$ so that it is consistent with the RGB image before passing it through the network. In our picking task, the camera is mounted on the top looking downward onto the objects in the tote. Therefore we set the height above ground to a constant value (1 in our case).

### B. Feature Embedding

The output segments from the COB can be very noisy varying from a small part of an object to a segment with multiple objects. Hence, it is important to learn a robust classifier that can classify these noisy segments into correct object categories. Traditional losses for classification such as softmax and binary cross entropy (BCE) do not suit the task at hand because these losses require the number of categories to be defined and fixed a-priori. To overcome this, we use a deep metric learning approach that has shown promising results on handling unseen categories [20], [1], [18]. As opposed to standard classifiers that learn the class specific information, the metric learning approach attempts to learn the general concept of distance metrics in the embedding space and thus can generalise well to unseen categories. In addition, a simple nearest neighbour classifier on the learned embedding space can be used to classify the given objects.

In this paper, we employ the deep metric learning approach proposed in [1] for learning the embedding space. This method learns a convolutional neural network (CNN) $F : \mathbb{R}^{n \times n} \to \mathbb{R}^d$ that directly maps the input image $x_i \in \mathbb{R}^{n \times n}$ to a low-dimensional embedding space $F(x_i) \in \mathbb{R}^d$ where the images belonging to the same category are mapped to nearby points and the images belonging to different categories are mapped far apart. To learn such a mapping, the authors employ a Triplet Network that consists of three identical branches of CNNs that share the same weights as shown in Fig. 4. The input data for the Triplet Network consists of triplets of samples where each tiplet includes an anchor $x^a$, a positive $x^p$ and a negative $x^n$ such that $x^a$ and $x^p$ belong to the same category and $x^n$ belongs to a different category. The network is trained using a triplet loss $J^t$ that aims to separate the similar pairs $(x_i^a, x_i^p)$ from the dissimilar pairs $(x_i^a, x_i^n)$ by a margin $m$ [33]:

$$J^t(\mathbf{x}_i^a, \mathbf{x}_i^p, \mathbf{x}_i^n) = \max\left(0, 1 - \frac{\|F(\mathbf{x}_i^a) - F(\mathbf{x}_i^n)\|_2}{\|F(\mathbf{x}_i^a) - F(\mathbf{x}_i^p)\|_2 + m}\right). \quad (1)$$

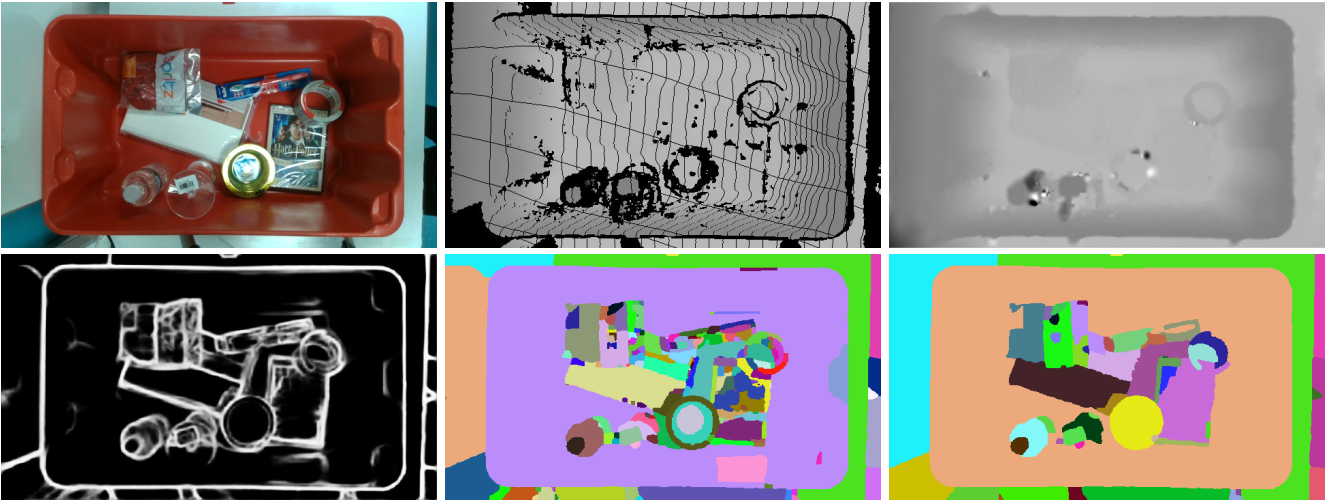

Fig. 2. An example of semantic-agnostic object segmentation. Top row: input color, raw depth and inpainted depth images. Bottom row: predicted boundary map and segmentations at two different scales. Different colours encode different object candidate regions, rather than semantic classes.

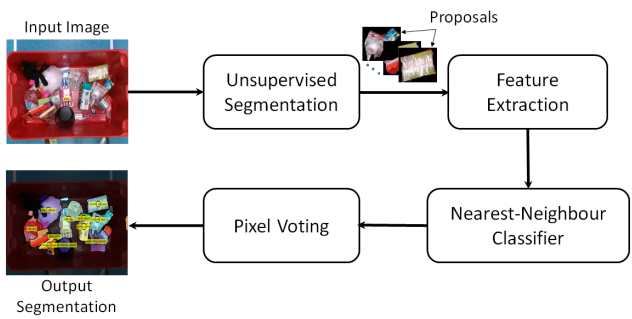

Fig. 3. Overview of the Deep Metric Learning approach.

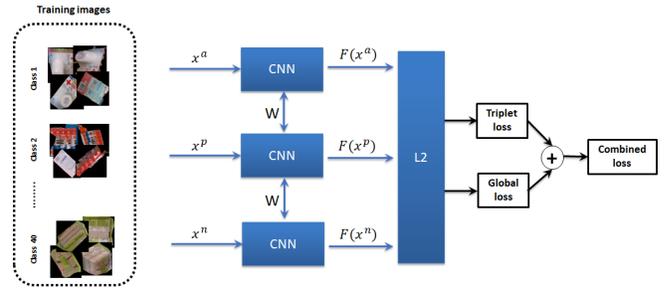

Fig. 4. A Triplet Network with a global loss. The figure is adopted from [1]. It is important to point out that the network is trained on images of *seen* categories only.

However, it was shown in [19], [1] that the triplet network ignores global structure of the embedding space and can lead to sub-optimal results. To overcome this, a global loss is proposed in [19], [1] that uses the statistics of the samples in the batch to improve the embedding. Specifically, it assumes that the distances between the similar pairs $d_i^+ = \|F(\mathbf{x}_i^a) - F(\mathbf{x}_i^p)\|_2^2/4$ and dis-similar pairs $d_i^- = \|F(\mathbf{x}_i^a) - F(\mathbf{x}_i^n)\|_2^2/4$

follow a distribution and the objective is to minimise the overlap $J^g$ between the two distributions:

$$J^g(\{\mathbf{x}_i^a, \mathbf{x}_i^p, \mathbf{x}_i^n\}_{i=1}^N) = \\ (\sigma^{2+} + \sigma^{2-}) + \lambda \max\left(0, \mu^+ - \mu^- + t\right), \quad (2)$$

where N is number of samples in the mini-batch, $t$ is the margin that separates the mean of the two distributions, In this paper, we use a weighted combination of the triplet and the global loss defined by:

$$J^c(\{\mathbf{x}_i^a, \mathbf{x}_i^p, \mathbf{x}_i^n\}_{i=1}^N) = \\ J^g(\{\mathbf{x}_i^a, \mathbf{x}_i^p, \mathbf{x}_i^n\}_{i=1}^N) + \alpha \sum_{i=1}^N J^t(\mathbf{x}_i^a, \mathbf{x}_i^p, \mathbf{x}_i^n), \quad (3)$$

where $\alpha$ is set to $0.8$ in our experiments.

**Implementation and training details:** For training the feature embedding model, we initialize the network with ImageNet pre-trained GoogLeNet [34] weights and randomly initialize the final fully connected layer similar to [18]. The learning rate for the randomly initialized fully connected layer is multiplied by 10 to achieve faster convergence. We use the images from our dataset consisting of 42 categories (40 training categories + $tote$ + $unlabelled$ ) to generate $800K$ triplets and train the network for 20 epochs. The images are first resized to $256 \times 256$ and then randomly cropped to $227 \times 227$. Similar to [1], we set the margin for the triplet and global loss to 0.2 and 0.01 respectively. We start experiment with an initial learning rate of 0.1 and gradually decrease it by a factor of 2 after every 3 epochs. We use a weight decay of 0.0005 for all of our experiments.

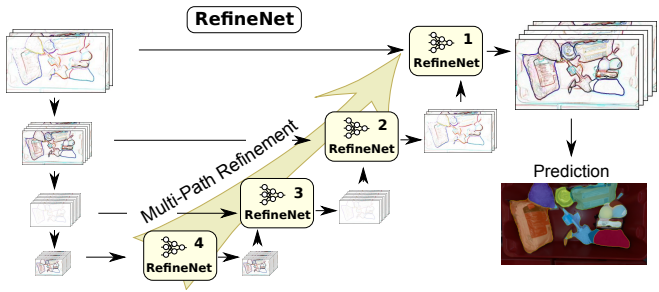

Fig. 5. A schematic overview of RefineNet adapted from [4]. ResNet [35] feature maps of varying resolutions are gradually combined to arrive at a refined high-resolution pixel-wise prediction.

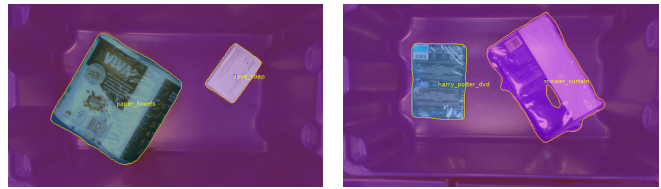

Fig. 6. Exemplar results of our automatic segmentation. A failure example is shown on the right, where the red part of the item is erroneously considered as background. Such failure cases are corrected manually by a human operator.

### C. Pixel Voting

As described above, the COB network produces an entire hierarchy of image segmentations for each image. Consequently, each pixel can be part of multiple segments. To resolve this ambiguity, a pixel voting scheme is used in the deep metric learning pipeline as a method to concatenate multiple labels for multiple segmentation proposals into a single layered segmentation map. In particular, the COB and feature embedding steps produce approximately 100 binary segmentation masks that are each assigned the $k$ nearest labels (we used $k = 3$ in our experiments). Because many segmentation masks overlap and the $k$ nearest labels may not belong to the same class, multiple labels are associated with the same pixels. The pixel voting method iterates over each label for every mask and accumulates a pixel-wise tally of how many times a class is associated with a pixel. Then, a list of expected classes, maintained by the robot system, is used to remove tallies for classes that are known to be absent from the image. Finally, for each pixel, the class with the highest tally is used as the label for that pixel. The output after pixel voting is a single segmentation map that can be directly used by the robot for manipulating a particular object.

## V. FULLY-SUPERVISED SEMANTIC SEGMENTATION

The above approach conceptually fits well into the ARC 2017 competition rules. It can segment the objects without any notion of semantics and it does not require any training to handle novel objects. However, it also has two major drawbacks. First, the RGB-D boundary detection is rather slow even on modern GPUs due to the computation of additional geometric features (i.e., HHA features) before being passed to the boundary network. Moreover, it requires an inpainting procedure to produce dense depth maps. Second, and perhaps more important, the entire segmentation pipeline consists of multiple sequential steps. Consequently, an error made in the boundary detection cannot be corrected later on. To remedy these shortcomings, we explore a second scene understanding approach based on semantic segmentation which can be trained end-to-end.

To that end, we adopt the recently developed RefineNet architecture [4] for our purpose. In a nutshell, RefineNet is a deep (fully) convolutional neural network (CNN), which exploits feature maps at different levels of detail to produce high-resolution semantic maps. A high-level overview is illustrated in Fig. 5. Compared to previous approaches that tackled high-resolution semantic segmentation such as DeepLab [22], RefineNet reduces the memory consumption and yields more accurate results on a number of public benchmarks. Its key idea is the adaptation of the identity mapping paradigm [36], which allows for effective gradient flow and consequently effective and efficient end-to-end training of very deep networks.

### A. Fast Data Collection

The 2017 Amazon Robotics Challenge (ARC) required robots to pick from a set of 50% known and 50% unknown items. The unknown item set was provided 45 minutes before an official run and available for the first 30 minutes of that time. The challenge here is that standard data collection and annotation approaches (like manual annotation of cluttered scenes) are infeasible to perform in the available time.

As a compromise to cluttered scenes, we opted to capture images of each new item without clutter, but with as many other commonalities to the final environment as possible. To achieve this, each item was placed in the Amazon-provided tote with the camera mounted on the robot's wrist at the same height above the scene as during a run. Each item was manually cycled through a number of positions and orientations to capture some of the variations the network would need to handle. As further described in Section VI-C, we chose to capture seven images of each unseen item. Note that we also experimented with a turntable solution. However, we found that manually placing and manipulating the items within the actual scene, i.e. tote or storage system, to be both more efficient and to yield more reliable training data for our purpose.

To speed up the annotation of these images, we employ the same RefineNet architecture as outlined above, but trained on only two classes for binary foreground/background segmentation. After each image is captured, the network outputs two segments for background and foreground, providing a segment of the item in the scene. To further accelerate the data collection task, we parallelise this approach by placing two items in the tote at a time such that they do not overlap in the image. When successful, the foreground mask contains two segments horizontally adjacent to one another (cf. Fig.6) that can be easily separated into two connected components. Labels are automatically assigned based on the assumption that the items in the scene match those read out by a human operator with an item list.

During the data capture process, another human operator visually verified each segment and class label, manually correcting any that are unsatisfactory for direct addition to the training set. After a few practice runs, a team of four members are able to capture 7 images of 16 items in approximately 4 minutes. An additional 3 minutes are taken by one human to finalise the manual check and correction procedure.

### B. Implementation and training details

Training is performed in two steps. We first train the model on our dataset with 41 training categories (40 objects and one background class), for 100 epochs, initialised with pre-trained ResNet-101 ImageNet weights. Note that the final softmax layer contains 16 (or 10 for the stow task) placeholder entries for unseen categories. Upon collecting the data as described above, we fine-tune the model using *all* available training data for 20 epochs within the available time frame. We use four NVIDIA GTX 1080Ti GPUs for training. Batch size 1 and learning rate $1e^{-4}$ is used for the initial fine-tuning stage, batch size 32 and learning rate $1e^{-5}$ is used for the final fine-tuning stage. It is important to note that we also exploit the available information about the presence or absence of items in the scene. The final prediction is taken as argmax not over all 57 classes, but only over the set of categories that are assumed to be present.

## VI. EXPERIMENTS

### A. Dataset

The 2017 contest challenged the teams by operating on a training and a competition item set. The former contained 40 items physically provided by Amazon Robotics to each team three months before the event, while the latter was revealed just 45 minutes before each run. Using the provided 40 training items and a curated set of 16 items to simulate the competition set, we produced a dataset to benchmark various vision systems. The unseen items set was selected to reflect the properties of training items and consisted mainly of typical household objects found in stores across Australia.

The dataset is composed of a train and test set, the training set is split into seen and unseen sets. The seen items training set comprises 137 images that contain between 0 and 20 items. Each of the 40 items in the seen items train set appear between 11 and 76 times. The unseen items training set comprises 120 images that contain 2 items each. Each of the 16 items appear 15 times. The test set comprises 67 images that contain between 2 and 20 items with an approximately equal split of seen and unseen items in each image. The dataset was captured using the hardware setup described in Sec. III and contains per-pixel labelled RGB images alongside aligned depth images.

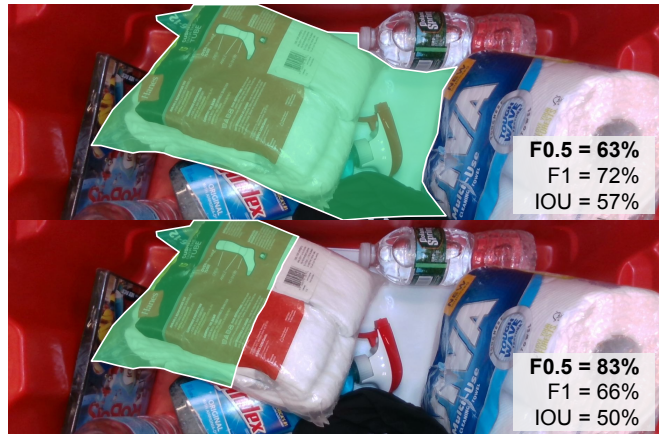

Fig. 7. An example illustrating the importance of different measures for grasping applications. **Top**: The object of interest, here, the tube socks, is undersegmented and the robot may pick a wrong item that is contained in the segmented region. **Bottom**: Only a part of the entire object is segmented correctly, yielding lower F1 and intersection-over-union (IOU) measures. It is evident, however, that this segmentation is more suitable for determining a correct grasp point and successfully manipulate the object. We argue that the $F_{0.5}$ measure is far more informative than the common IOU or $F_1$. Note that precision would also be indicative of success in this example, but should not be used in isolation because it loses information about recall.

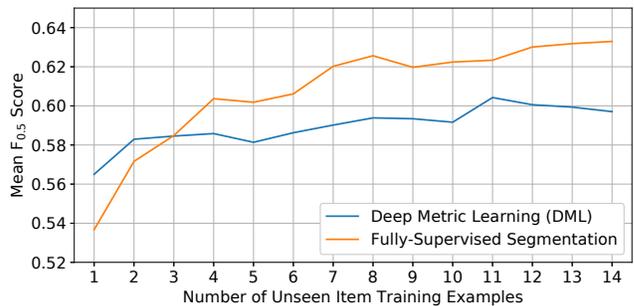

Fig. 8. We report the $F_{0.5}$ score of both fully-supervised segmentation (RefineNet) and our deep metric learning approach with respect to the number of unseen images used for training. We find that both improve with the number of images but see a significant difference in absolute performance between the two. Interestingly, DML clearly outperforms the fully-supervised CNN for one training example, but does not quite reach the performance when enough training data are available.

### B. Evaluation Criteria

The most common approaches for evaluating semantic segmentation vision systems are Intersection over Union (IOU) and $F_1$. We argue that these metrics are sub-optimal when benchmarking semantic segmentation vision systems for use in robotic applications.

Within the context of our robot, Cartman, the semantic segmentation system links a raw image of the scene to a grasp point selection system. Our grasp point selection system assumes that every pixel of a provided segment is the target item. Grasp points are generated using heuristics such as distance from segment edge and direction of surface normals. With our grasp point selection system in mind, receiving a segment that contains the pixels of neighbouring items may result in the robot completely grasping the wrong item. As depicted in Fig. 7, the measures of IOU and $F_1$ do not capture the desirability of a given segment for *this* type of grasp selection system. This point highlights that the choice of metric is critical and provides our justification for why the $F_{0.5}$ measure, defined as

$$F_{0.5} = 1.25 * \frac{\text{precision} \cdot \text{recall}}{0.25 \cdot \text{precision} + \text{recall}} \quad (4)$$

is used throughout this paper. $F_{0.5}$ differs from $F_1$ by weighting recall less than precision, better differentiating desirable and undesirable item segments.

*C. Results*

We report results for both deep metric learning (DML) and fully-supervised semantic segmentation on 67 cluttered scenes with an approximately even mix of seen and unseen items. As shown in Fig. 8, both approaches improve with an increasing number of unseen item images. Our experiments showed that the main bottleneck for the underwhelming performance of DML is the COB step, which provides rather noisy object segmentation. Using perfect segment, DML achieves an $F_{0.5}$ score of about 0.85. However, with the fully-supervised approach being both more efficient and more accurate for the relevant case of multiple training examples, it was selected for use in the final competition system.

We capture seven images of each item from the competition set before each official run to finetune our semantic segmentation approach. Seven images were found to provide the perfect balance between time to capture the data and overall performance, as shown in Fig. 9. While the time frame is sufficiently large to collect 12 images per item and gain a slight improvement in performance, we prefer to have a safety buffer for re-collecting the data and re-training the model for a second time in case if something goes wrong. All following experiments are conducted using the fine-tuned RefineNet model on seven images per unseen object.

We perform two additional tests that help to characterise the semantic segmentation network. Firstly, Fig. 10 shows how the number of appearances of an item in the training set influences the $F_{0.5}$ score of that item across the test set. We find no correlation between the two, indicating that performance is rather a function of an item's appearance, and not necessarily how many training examples of the item are available. Secondly, we analyse in Fig. 11 how the number of objects in a scene impacts the mean $F_{0.5}$ score of the network on each image in our test set. As expected, the performance semantic segmentation degrades monotonically as more and more objects are present and the scene becomes more cluttered.

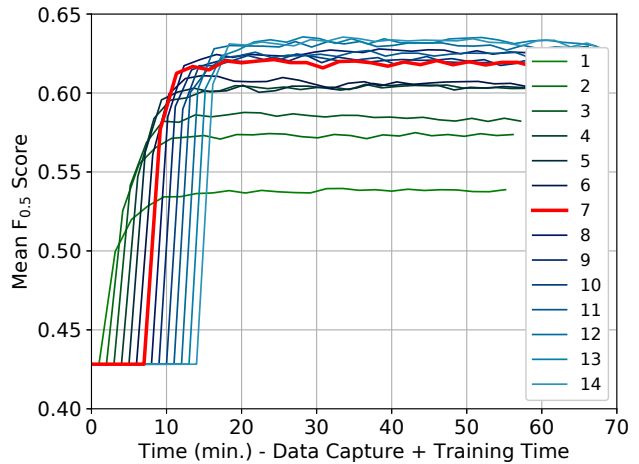

Fig. 9. Each line represents the entire time needed to collect $n$ images for all unseen items and train a model to reach a certain $F_{0.5}$ score. Our operating mode of training with 7 images for each item is highlighted in bold red [25].

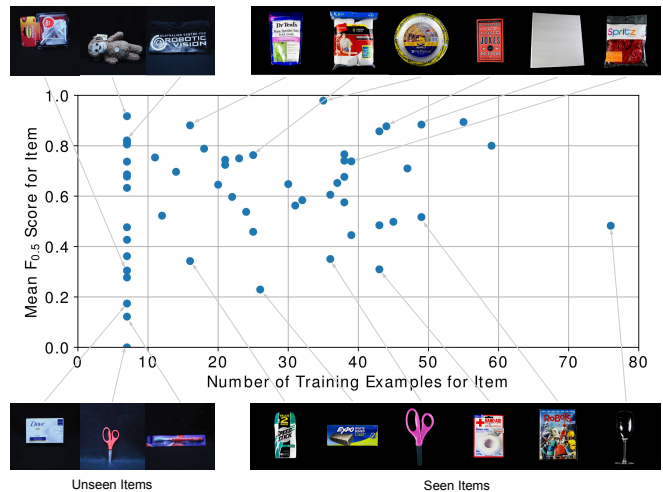

Fig. 10. A detailed analysis of per-class performance as a function of the number of available training samples. See text for details.

Table I provides a quantitative overview of the two methods discussed above. Even though our chosen approach (FSS) yields higher accuracy, the DML method is a better candidate when fewer training images or less time for data collection are available. Note that most of the time spent on DML is actually due to depth inpainting and HHA feature computation for the boundary detection step. The classification alone is very efficient because it consists of extracting features and finding a nearest neighbour.

TABLE I
QUANTITATIVE COMPARISON OF OUR APPROACHES.

| Method | Input | Training | | Prediction | $F_{0.5}$ | |
|---|---|---|---|---|---|---|
| | | Offl. | Onl. | | 1 img. | 7 imgs. |
| DML | RGB-D | 24h | **0s** | 8s | **0.57** | 0.59 |
| FSS | RGB | **1h** | 10m | **0.2s** | 0.53 | **0.62** |

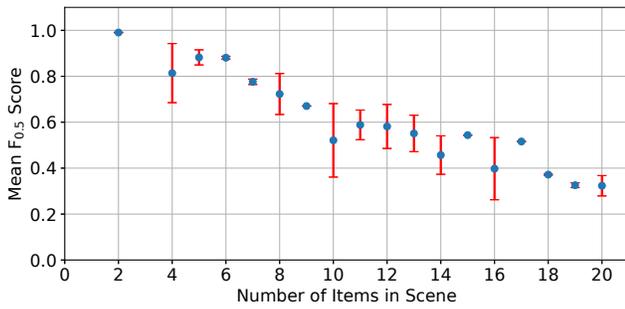

Fig. 11. We report how the number of items in a scene changes the $F_{0.5}$ score of RefineNet on our test set.

## VII. CONCLUSIONS

We presented two segmentation approaches that were developed to win the 2017 Amazon Robotics Challenge. One following a deep metric learning strategy without the need to retrain the model to handle new object categories, the other a state-of-the-art semantic segmentation CNN which can be fine-tuned using very few training examples. To obtain the training data, we also introduced an effective data collection scheme. Finally, we investigated the importance of segmentation measures in the context of robotic manipulation.